\documentclass[11pt,letterpaper]{article}
\usepackage{emnlp2017}
\usepackage{times}
\usepackage{latexsym}
\usepackage{microtype}
\usepackage{latexsym}
\usepackage{multirow}
\usepackage{subfigure}
\usepackage{graphicx,subfigure}
\usepackage{amssymb}
\usepackage[hang,flushmargin]{footmisc}
\usepackage{url}

\emnlpfinalcopy

\def\emnlppaperid{***}

\title{Lexicon Integrated CNN Models with Attention for Sentiment Analysis}

\author{Bonggun Shin, Timothy Lee, Jinho D. Choi \\
Math and Computer Science \\
Emory University \\
Atlanta, GA 30322 \\
{\tt {\{bonggun.shin,timothy.lee,jinho.choi\}}@emory.edu} \\}

\date{}
\hypersetup{draft}

\begin{document}

\maketitle

\begin{abstract}
  With the advent of word embeddings, lexicons are no longer fully utilized for sentiment analysis although they still provide important features in the traditional setting. This paper introduces a novel approach to sentiment analysis that integrates lexicon embeddings and an attention mechanism into Convolutional Neural Networks. Our approach performs separate convolutions for word and lexicon embeddings and provides a global view of the document using attention. Our models are experimented on both the SemEval'16 Task 4 dataset and the Stanford Sentiment Treebank and show comparative or better results against the existing state-of-the-art systems. Our analysis shows that lexicon embeddings allow building high-performing models with much smaller word embeddings, and the attention mechanism effectively dims out noisy words for sentiment analysis.
\end{abstract}

\section{Introduction}
\label{sec:introduction}

Sentiment analysis is a task of identifying sentiment polarities expressed in documents, typically positive, neutral, or negative.
Although the task of sentiment analysis has been well-explored~\cite{mullen2004sentiment,pang2005seeing,wilson2005recognizing}, it is still very challenging due to the complexity of extracting human emotion from raw text.
The recent advance of deep learning has definitely elevated the performance of this task~\cite{socher2013recursive,kim2014convolutional,yin2016multichannel} although there is much more room to improve.


In the traditional setting where statistical models are based on sparse features, lexicons consisting of words and their sentiment scores are shown to be highly effective for sentiment analysis because they provide features that may not be captured from training data~\cite{hu2004mining,kim2004determining,ding2008holistic,taboada2011lexicon}.
However, since the appearance of word embeddings, the use of lexicons is getting faded away because word embeddings are believed to capture the sentiment aspects of those words.
This brought us two important questions:

\begin{itemize}
\item Can lexicons be still useful for sentiment analysis when coupled with word embeddings?
\item If yes, what is the most effective way of incorporating lexicons with word embeddings?
\end{itemize}

\noindent To answer these questions, we first construct lexicon embeddings that are specifically designed for sentiment analysis and integrate them into the existing Convolutional Neural Network (CNN) model similar to \newcite{kim2014convolutional}.
Three ways of lexicon integration to the CNN model are proposed, which show distinctive characteristics for different genres (Section~\ref{ssec:lexcion-integration}).
We then incorporate an efficient attention mechanism to our CNN models, which provides a global view of the document by emphasizing (or de-emphasizing) important words and lexicons (Section~\ref{ssec:embedding-attention}).
Our models using lexicon embeddings are evaluated on two well-known datasets, the SemEval'16 dataset and the Stanford Sentiment Treebank, and show state-of-the-art results on both datasets (Section~\ref{sec:experiments}).
To the best of our knowledge, this is the first time that lexicon embeddings are introduced for sentiment analysis.

\section{Related Work}
\label{sec:related-work}

The first attempt of sentiment analysis on text was initiated by \newcite{pang2002thumbs} who pioneered this field by using bag-of-word features.
This work mostly hinged on feature engineering; since then, many kinds of feature learning methods had been introduced to increase the performance 
~\cite{pang2008opinion,liu2012sentiment,gimpel2011part,feldman2013techniques,mohammad2013nrc}.
Aside from pure machine learning approaches, lexicon based approaches had been another trend, which relied on the manual or algorithmic creation of word sentiment scores~\cite{hu2004mining,kim2004determining,ding2008holistic,taboada2011lexicon}.

Since the emergence of the Convolutional Neural Networks (CNN; \newcite{collobert2011natural}),
conventional methods have become gradually obsolete because of the outstanding performance from the CNN variants.
CNN based models are distinguished from earlier methods because they do not rely on laborious feature engineering.
The first success of CNN in sentiment analysis was triggered by document classification research~\cite{kim2014convolutional}, where CNN showed state-of-the-art results in numerous document classification datasets.
This success has engendered an upsurge in deep neural network research for sentiment analysis.
Various modified models have been proposed in the literature.
One of the famous deep learning methods that models a document is the generalized phrase proposed by \newcite{yin2014exploration}, which represents a sentence using element-wise addition, multiplication, or recursive auto-encoder.

Endeavors to capture $n$-gram information bore fruits with CNN, max pooling, and softmax~\cite{collobert2011natural,kim2014convolutional}, which is regarded as the standard methods of the document classification problem these days.
\newcite{kalchbrenner2014convolutional} extended this standard CNN model with dynamic k-max pooling, which served as an input layer to another stacked convolution layer.
Multichannel CNN methods~\cite{kim2014convolutional,yin2016multichannel} are another branch of CNN, 
where assorted embeddings are considered together when convolving the input.
Unlike \newcite{kim2014convolutional}'s model that relies on a single type of embedding with different mutability characteristics 
of the weights of embedding layer, \newcite{yin2016multichannel} incorporates diverse sort of embedding types using multichannel CNN.

Two notable pioneers in using lexicon for sentiment analysis are \newcite{MohammadKZ2013,kalchbrenner:2014} generated scores with other manually generated sentiment lexicon scores to achieved the state-of-the-art result in SemEval-2013 Twitter sentiment analysis task. In general domain, \newcite{hu2004mining} generated a user review lexicon that showed promising result in capturing sentiment in customer product reviews.

\noindent Attention based methods have been successful in many application domains, such as image classification~\cite{stollenga2014deep}, image caption generation~\cite{xu2015show}, machine translation~\cite{cho2014learning,bahdanau2014neural,luong2015effective}, and question answering~\cite{shih2015look,chen2015abc,yang2015stacked}.
However, in the field of sentiment analysis, the attention is applied to only aspect-based sentiment classification~\cite{yanase2016bunji}.
To the best knowledge of ours, there is no attention-based model for a general sentiment analysis task.





\section{Approach}
\label{sec:approach}

The models proposed here are based on a convolutional architecture and use naive concatenation (Section~\ref{sssec:navie-concatenation}), multichannel (Section~\ref{sssec:multichannel}), separate convolution (Section~\ref{sssec:separate-convolution}), and embedding attention (Section~\ref{ssec:embedding-attention}) for the integration of lexicon embeddings to CNN.

\subsection{Baseline}
\label{ssec:baseline}

Our baseline approach is a one-layer CNN model using pre-trained word embeddings, which is a reimplementation of the CNN model introduced by \newcite{kim2014convolutional}. 
Let $s \in \mathbb{R}^{n \times d}$ be a matrix representing the input document, where $n$ is the number of words, $d$ is the dimension of the word embeddings, and each row corresponds to the word embedding, $w_i \in \mathbb{R}^d$, where $w_i$ indicates the $i$'th word in the document. 
This document matrix $s$ is fed into the convolutional layer and convolved by the weights $c\in \mathbb{R}^{l \times d}$, where $l$ is the length of the filter.

The convolutional layer can take $m$-number of filters of the length $l$.
Each convolution produces a vector $v_c \in \mathbb{R}^{n-l+1}$, where elements in $v_c$ convey the $l$-gram features across the document.
The max pooling layer selects the most salient features from each of the $m$ vectors produced by the filters. 
As a result, the output of this max pooling layer is a vector $v_m \in \mathbb{R}^{(n-l+1) \times m}$. 
The selected features are passed onto the softmax layer, which is optimized for the score of each sentiment class label.

\subsection{Lexicon Integration}
\label{ssec:lexcion-integration}

Lexicon embeddings are derived by taking scores from multiple sources of lexicon datasets.
Each lexicon dataset consists of key-value pairs, where the key is a word and the value is a list of sentiment scores for that word (e.g., probabilities of the word in positive, neutral, and negative contexts).
These scores range between $-1$ and $1$, where $-1$ and $1$ being the most negative and positive, respectively.
However, some lexicons contain non-probabilistic scores (e.g., frequency counts of the word in sentimental contexts), which are normalized to $[-1,1]$.

\begin{figure}[ht!]
\centering     
\subfigure[Naive concatenation (Section~\ref{sssec:navie-concatenation}). The lexicon embeddings (on the right) are concatenated to the word embeddings (on the left).]
{\label{fig:nc}\includegraphics[width=70mm]{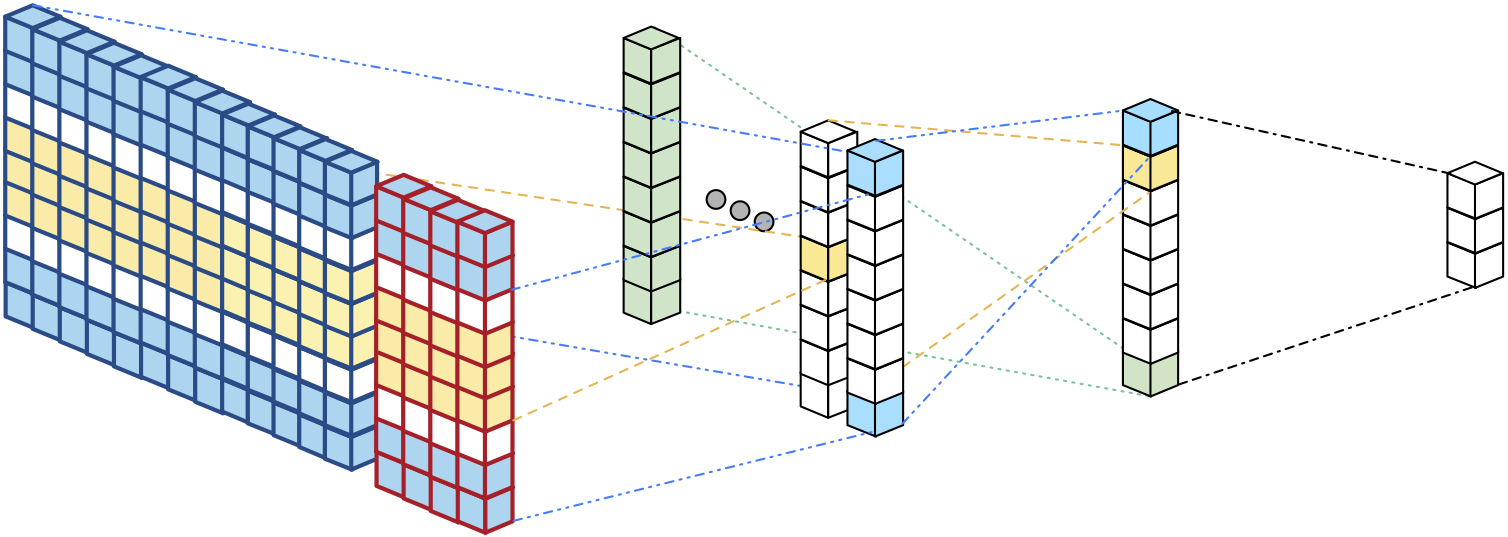}}
\subfigure[Multichannel (Section~\ref{sssec:multichannel}). The lexicon embeddings are added to the second channel whereas the word embeddings are added to the first channel.]
{\label{fig:mc}\includegraphics[width=70mm]{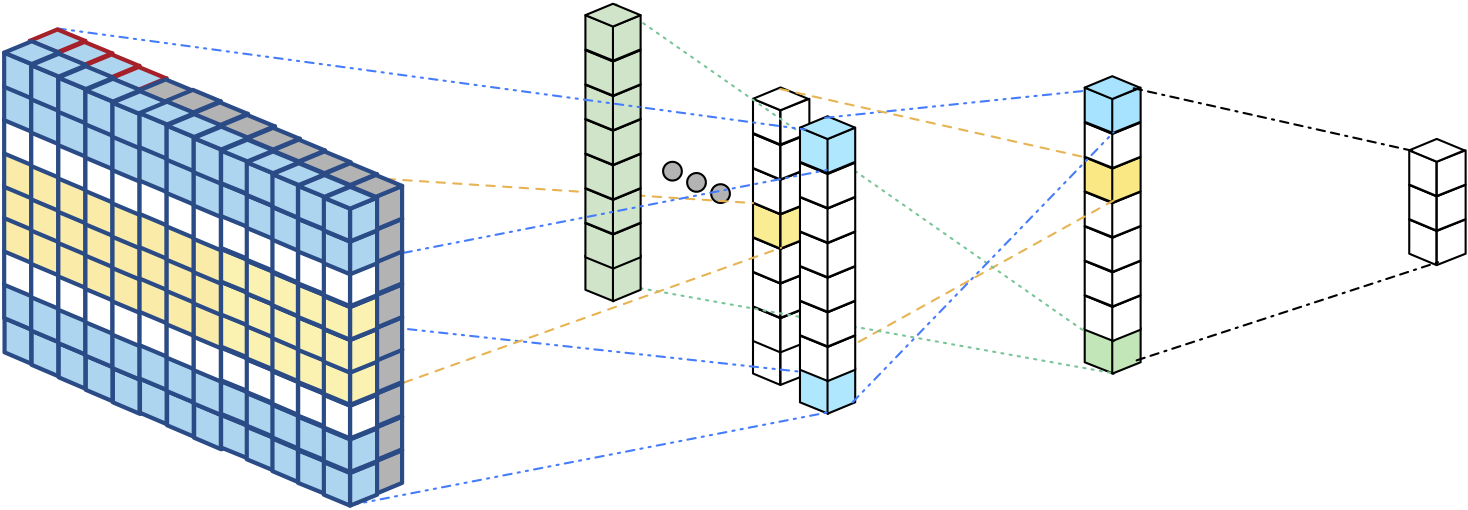}}
\subfigure[Separate convolution (Section~\ref{sssec:separate-convolution}). The lexicon embeddings are processed by a separate convolution (on the right) from the word embeddings (on the left).]
{\label{fig:sc}\includegraphics[width=70mm]{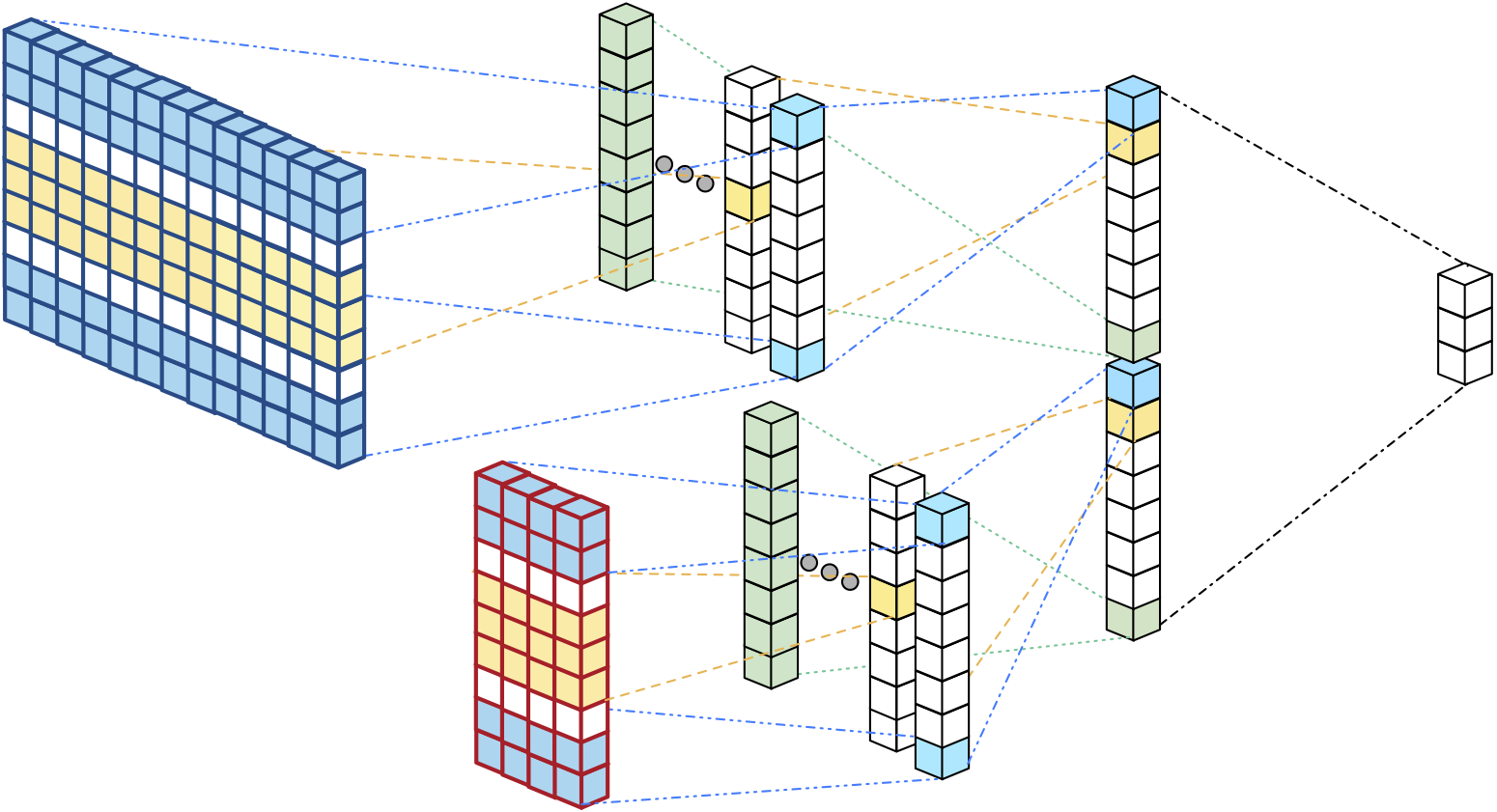}}
\caption{Lexicon integration to the CNN model.}
\end{figure}

\noindent For each word $w \in W$, where $W$ is the union of all words in the lexicon datasets, a lexicon embedding is constructed by concatenating all the scores among the datasets with respect to $w$.
If $w$ does not appear in certain datasets, $0$ values are assigned in place. 
The resulting embedding is in the form of a vector $v \in \mathbb{R}^{e}$, where $e$ is the total number of scores across all lexicon datasets.
The following subsections propose three methods for lexicon integration to the baseline CNN model (Section~\ref{ssec:baseline}), which depict different characteristics depending on the peculiarities of each domain.


\subsubsection{Naive Concatenation}
\label{sssec:navie-concatenation}

The simplest way of blending a lexicon embedding into its corresponding word embedding is to append it
to the end of the word embedding (Figure~\ref{fig:nc}). 
In a formal notation, the document matrix becomes $s \in \mathbb{R}^{n \times (d + e)}$.
The subsequent process is the same as the baseline approach.

\subsubsection{Multichannel}
\label{sssec:multichannel}

Inspired by \newcite{yin2016multichannel} who integrated several kinds of word embeddings using multichannel CNN, lexicon embeddings in this approach are represented in another channel along with the word embedding channel where both channels are convolved together (Figure~\ref{fig:mc}).
Since the dimension of lexicon embeddings is considerably smaller than that of word embeddings (i.e., $d \gg e$), zeros are padded to the lexicon embeddings so their dimensions match (i.e., $d = e$).
The identical shape of these two channels allows multichannel convolution to the input document.


\subsubsection{Separate Convolution}
\label{sssec:separate-convolution}

Another way of adding lexicon embeddings to the CNN model is to process a separate convolution for them (Figure~\ref{fig:sc}).
In this case, two individual convolutions are applied to word embeddings and lexicon embeddings.
The max pooled output features from each convolution are then merged together to form an input vector to the softmax layer.
Formally, let $l_{w}$, $l_{x}$ be the filter lengths for word embeddings and lexicon embeddings, respectively.
Let $m_{w}$ and $m_{x}$ be the numbers of filters for word embeddings and lexicon embeddings, respectively.
The resulting penultimate layer includes max pooled features from word embeddings and lexicon embeddings of size $[(n - l_w + 1) \times m_w] + [(n - l_x + 1) \times m_x]$.

\subsection{Embedding Attention}
\label{ssec:embedding-attention}

\begin{figure}[ht!]
\centering     
\subfigure[Given a document matrix, the attention matrix is first created by performing multiple convolutions. The attention vector is then created by performing max pooling on each row of the attention matrix.]
{\label{fig:eav1}\includegraphics[trim={0 0ex 0 0},scale=0.3]{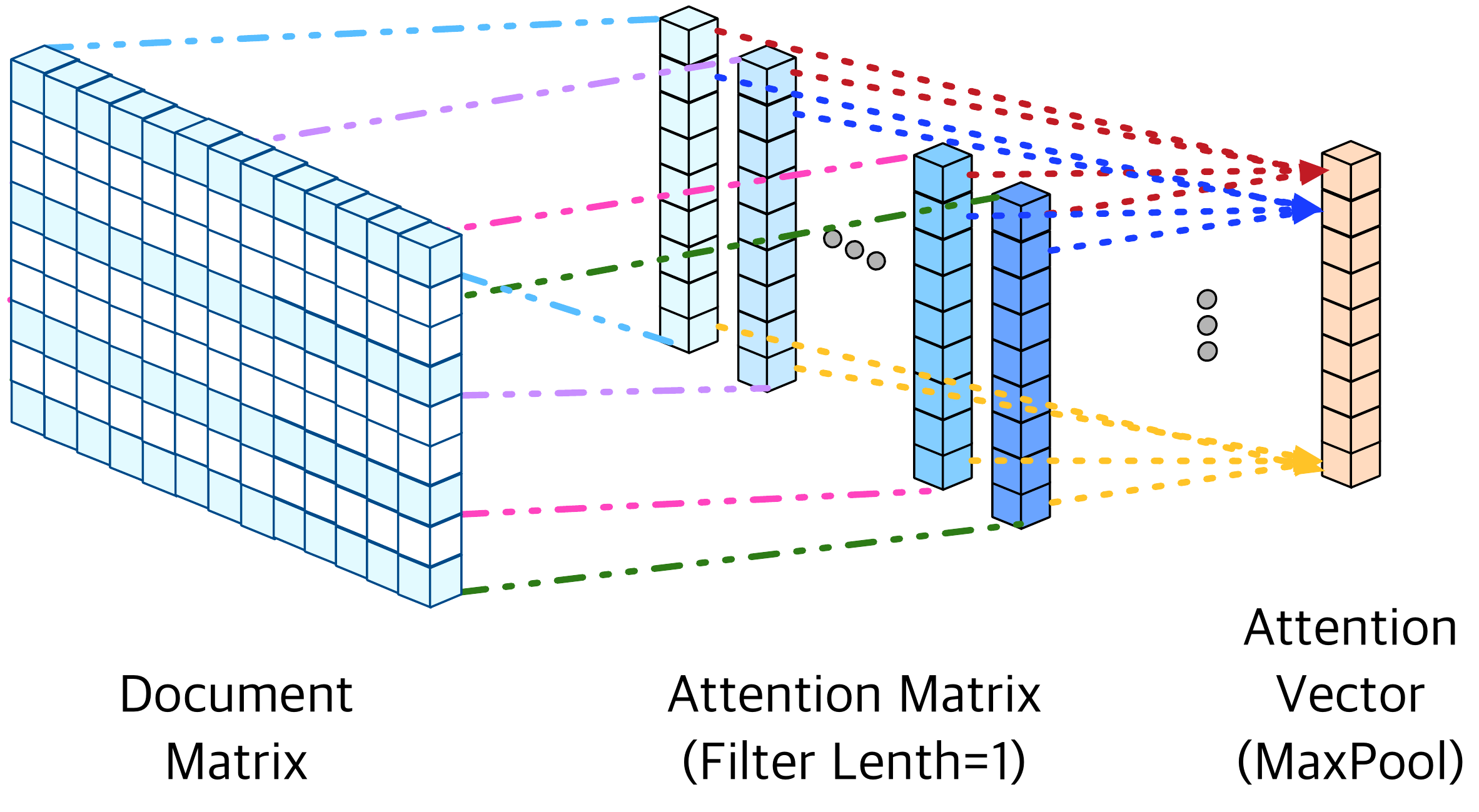}}
\subfigure[The embedding attention vector is created by multiplying the transposed document matrix to the attention vector.]
{\label{fig:eav2}\includegraphics[trim={-15ex 0ex -15ex 0},scale=0.3]{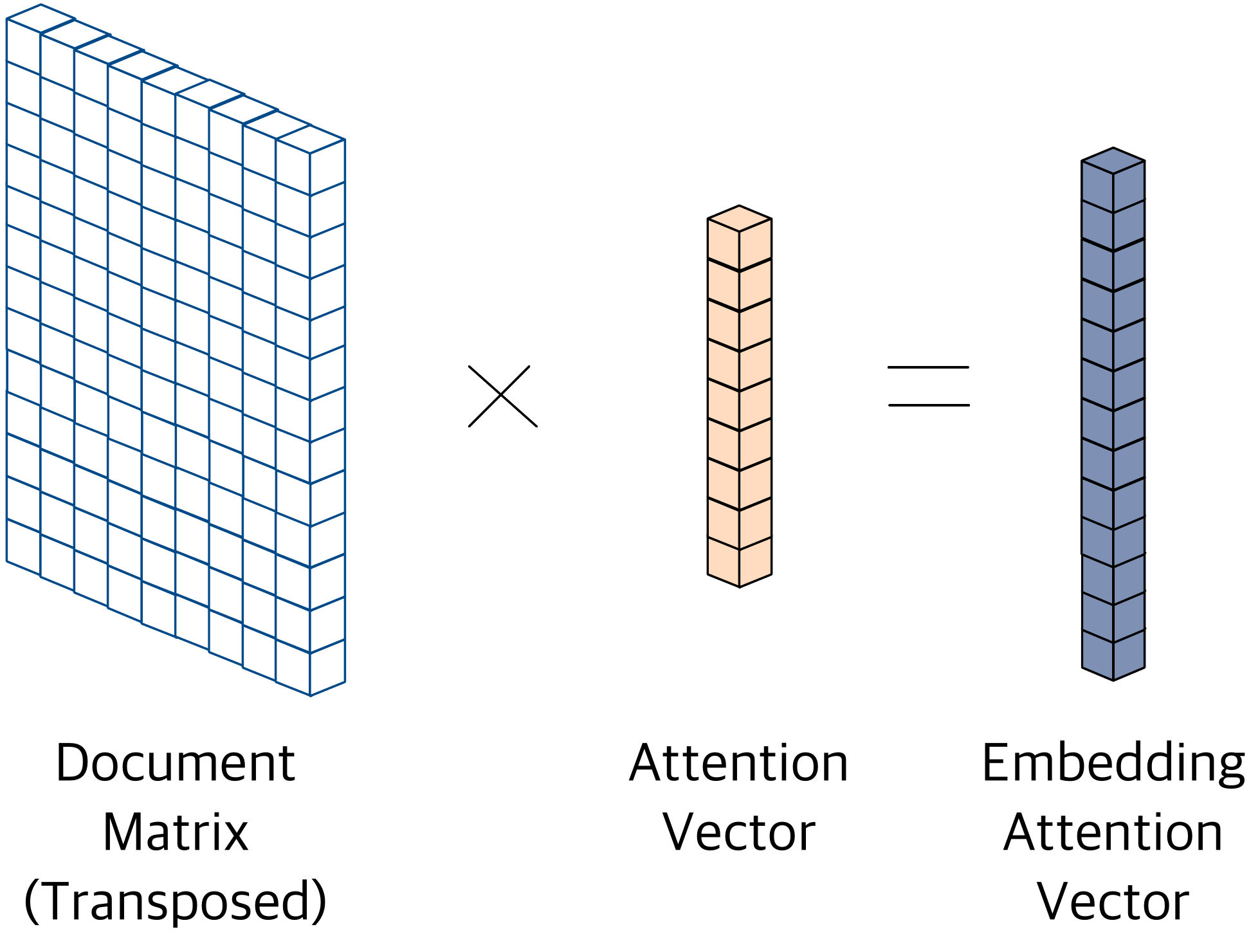}}
\caption{Construction of the embedding attention vector from a document matrix.}
\label{fig:embedding-attention}
\end{figure}

Section~\ref{ssec:lexcion-integration} describes how lexicon embeddings can be incorporated into the CNN model in Section~\ref{ssec:baseline}.
Each CNN model uses several filters with different lengths; given the filter length $l$, the convolution considers $l$-gram features.
However, these $l$-gram features account only for local views, not the global view of the document, which is necessary for several transitional cases such as negation in sentiment analysis~\cite{socher2012semantic}.
To ameliorate this issue, we introduce the embedding attention vector (EAV), which transforms the document matrix in each embedding space into a vector.
For example, the EAV in the word embedding space is calculated as a weighted sum of each column in the document matrix $s \in \mathbb{R}^{n \times d}$, which yields a vector $v \in \mathbb{R}^{d}$.
For each document, two EAVs can be derived, one from the document matrix consisting of word embeddings and the other from the one consisting of lexicon embeddings.
All embeddings in the document matrix are used to create the EAV through multiple convolutions and max pooling as follows:

\vspace{-0.5ex}
\begin{enumerate}
\item Apply $m$-number of convolutions with the filter length $1$ to the document matrix $s \in \mathbb{R}^{n \times d}$. For lexicon embeddings, the document matrix has a dimension of $\mathbb{R}^{n \times e}$.\vspace{-1.5ex}
\item Aggregate all convolution outputs to form an attention matrix $s_a \in \mathbb{R}^{n \times m}$, where $n$ is the number of words in the document, and $m$ is the number of filters whose length is $1$.\vspace{-1.5ex}
\item Execute max pooling for each row of the attention matrix $s_a$, which generates the attention vector $v_a \in \mathbb{R}^n$ (Figure~\ref{fig:eav1}).\vspace{-1.5ex}
\item Transpose the document matrix $s$ such that $s^T \in \mathbb{R}^{d \times n}$, and multiply it with the attention vector $v_a \in \mathbb{R}^{n}$, which generates the embedding attention vector $v_e \in \mathbb{R}^d$ (Figure~\ref{fig:eav2}).
\end{enumerate}
\vspace{-1.5ex}

\noindent The resulting EAVs are appended to the penultimate layer to serve as additional information for the softmax layer.
For our experiments, EAVs are generated from both word and lexicon embedding spaces for all of the three lexicon integration methods in Section~\ref{ssec:lexcion-integration}. 
 

\section{Experiments}
\label{sec:experiments}

\subsection{Corpora}

\subsubsection{SemEval-2016 Task 4}
\label{sssec:semeval-data}

All models are evaluated on the micro-blog dataset distributed by the SemEval'16 Task 4a~\cite{nakov-EtAl:2016:SemEval1}.
The dataset is gleaned from tweets with annotation of three sentiment classes: positive, neutral, and negative.
The available dataset contains only tweet IDs and their sentiment polarities; the actual tweet texts are not included in this dataset due to the copyright restrictions.
Although the download script provided by SemEval'16 gives a way of accessing the actual texts on the web, a portion of tweets is no longer accessible.
To compensate this loss, the dataset also includes tweet instances from the SemEval'13 task.


\begin{table}[htbp!]
\vspace{-1ex}
\centering\small
\begin{tabular}{l||r|r|r||r}
 & \multicolumn{1}{c|}{\texttt{\textbf{+}}} & \multicolumn{1}{c|}{\texttt{\textbf{0}}} & \multicolumn{1}{c||}{\texttt{\textbf{-}}} & \multicolumn{1}{c}{\bf All} \\
\hline\hline
\texttt{TRN} & 6,480 &  6,577 & 2,328 & 15,385 \\
\texttt{DEV} &   786 &    548 &   254 &  1,588 \\
\texttt{TST} & 7,059 & 10,342 & 3,231 & 20,632 \\ 
\end{tabular}
\caption{Statistics of the SemEval'16 Task 4 dataset. \texttt{+}/\texttt{0}/\texttt{-}: positive/neutral/negative, \texttt{TRN}/\texttt{DEV}/\texttt{TST}: training, development, evaluation sets.}
\label{tbl:semeval-data}
\end{table}

\noindent The classification results are evaluated by averaging the F1-scores of positive and negative sentiments as suggested by the SemEval'16 Task 4a.

\subsubsection{Stanford Sentiment Treebank}
\label{sssec:sst-data}

Another dataset consisting of movie reviews from Rotten Tomatoes is used for evaluating the robustness of our models across different genres.
This dataset, called the Stanford Sentiment Treebank, was originally collected by \newcite{pang2005seeing} and later extended by \newcite{socher2013recursive}.
The sentiment annotation in this dataset is categorized into five classes: very positive, positive, neutral, negative, and very negative.
Following the previous work~\cite{kim2014convolutional}, the results are evaluated by the conventional classification accuracy.

\begin{table}[htbp!]
\vspace{-1ex}
\centering\small
\begin{tabular}{l||r|r|r|r|r||r}
 & \multicolumn{1}{c|}{\bf ++} & \multicolumn{1}{c|}{\bf +} & \multicolumn{1}{c|}{\bf 0} & \multicolumn{1}{c|}{\bf -} & \multicolumn{1}{c||}{\bf --} & \multicolumn{1}{c}{\bf All} \\
\hline
\texttt{TRN} & 1288 & 2322 & 1624 & 2218 & 1092 & 8,544\\
\texttt{DEV} & 165 & 279 & 229 & 289 & 139 & 1,101\\
\texttt{TST} & 399 & 510 & 389 & 633 & 279 & 2,210\\ 
\end{tabular}
\caption{Statistics of the Stanford Sentiment Treebank dataset. ++/+/0/-/--: very positive/positive/ neutral/negative/very negative.}
\label{tbl:sst-data}
\end{table}

\subsection{Embedding Construction}

\subsubsection{Word Embeddings}

To best capture the word semantics in each genre, different corpora are used to train word embeddings for the SemEval'16 (S16) and the Stanford Sentiment Treebank (SST) datasets.
For S16, word embeddings are trained on tweets collected by the Archive Team,\footnote{\url{archive.org/details/twitterstream}} consisting of 3.67M word types.
For SST, word embeddings are trained on the Amazon Review dataset,\footnote{\url{snap.stanford.edu/data/web-Amazon.html}} containing 2.67M word types.

All documents are pre-tokenized by the open-source toolkit, NLP4J.\footnote{\url{github.com/emorynlp/nlp4j}}
The word embeddings are trained by the original implementation of word2vec from Google using skip-gram and negative sampling.\footnote{\url{code.google.com/p/word2vec}}
No explicit hyper-parameter tuning is performed.
For each genre, four sets of embeddings with different dimensions (50, 100, 200, 400) are trained to observe the impact of the embedding size on each approach. 


\subsubsection{Lexicon Embeddings}
\label{sssec:lexicon-embeddings}

Six types of sentiment lexicons are used to build lexicon embeddings.
All lexicons include sentiment scores; some lexicons contain information about the frequency of positive and negative sentiment polarity associated with each word:

\begin{itemize}
\item National Research Council Canada (NRC) Hashtag Affirmative and Negated Context Sentiment Lexicon~\cite{kiritchenko2014sentiment}.
\item NRC Hashtag Sentiment Lexicon\\\cite{MohammadKZ2013}.
\item NRC Sentiment140 Lexicon\\\cite{kiritchenko2014sentiment}.
\item Sentiment140 Lexicon\\\cite{MohammadKZ2013}.
\item MaxDiff Twitter Sentiment Lexicon\\\cite{kiritchenko2014sentiment}.
\item Bing Liu Opinion Lexicon\\\cite{hu2004mining}.
\end{itemize}


\noindent When creating lexicon embeddings, the narrow coverage of vocabulary in lexicons often raises missing scores. 
If a given word is missing in a specific lexicon, neutral scores of 0 are substituted.

\noindent Table~\ref{tbl:summaryStatisticEmbedding} shows the word type coverage of our word and lexicon embeddings for each dataset.
The lexicon embeddings show relatively poor coverage; nevertheless, our experiments show that these lexicon embeddings help sentiment classification in various ways (Section~\ref{ssec:evaluation}).

\begin{table}[htbp!]
\centering\small
\begin{tabular}{c||c|c||c|c}
 & \multicolumn{2}{c||}{\bf Word Emb} & \multicolumn{2}{c}{\bf Lexicon Emb} \\
\cline{2-5}
 & \bf S16 & \bf SST & \bf S16 & \bf SST \\
\hline\hline
\texttt{TRN} & 70.12 & 97.66 & 11.53 & 9.20 \\
\texttt{DEV} & 81.90 & 98.91 &  3.29 & 3.32 \\
\texttt{TST} & 68.57 & 98.58 & 12.40 & 4.98 \\
\end{tabular}
\caption{The percentage of word types covered by our word and lexicon embeddings for each dataset.}
\label{tbl:summaryStatisticEmbedding}
\end{table}

\subsection{Evaluation}
\label{ssec:evaluation}

\begin{table*}[htp!]
\centering\small
\begin{tabular}{c||c|c}
\bf Model & S16 (Avg F1 Score) & SST (Accuracy) \\
\hline \hline
Baseline                       &         61.6  &         47.5 \\
\hline
NC                             &         63.4  &         46.8 \\
MC                             &         61.8  &         47.0 \\
SC                             &         63.6  &         47.5 \\
\hline
NC-EAV                         &         63.4  & \textbf{48.8}\\
MC-EAV                         &         62.1  &         47.3 \\
SC-EAV                         & \textbf{63.8} & \textbf{48.8}\\
\hline\hline
\newcite{deriu2016swisscheese}       & 63.3    & - \\
\newcite{rouvier-favre:2016:SemEval} & 63.0    & - \\
\hline
\newcite{kim2014convolutional}       & -       & 48.0 \\
\newcite{kalchbrenner:2014}          & -       & 48.5 \\
\newcite{DBLP:conf/icml/LeM14}       & -       & 48.7 \\
\newcite{yin2016multichannel}$^*$    & -       & \bf 49.6 \\
\end{tabular}
\caption{Evaluation set results (random seed is fixed to 1) of the proposed models in comparison to the state-of-the-art approaches.
\label{tbl:resultcomparison}
\textbf{\newcite{deriu2016swisscheese}}: the first place for the SemEval'16 task 4a using an ensemble of two CNN models.
\textbf{\newcite{rouvier-favre:2016:SemEval}}: the second place for the SemEval'16 task 4a using various embeddings in CNN. 
\textbf{\newcite{kim2014convolutional}}: the state of the art single layer CNN model.
\textbf{\newcite{kalchbrenner:2014}}: dynamic CNN with k-max pooling.
\textbf{\newcite{DBLP:conf/icml/LeM14}}: logistic regression on top of paragraph vectors.
\textbf{\newcite{yin2016multichannel}}: the state-of-the-art dual layer CNN with five channel embeddings. 
}
\end{table*}

Seven models are evaluated to show the effectiveness of lexicon embeddings to sentiment analysis: baseline (Section~\ref{ssec:baseline}), naive concatenation (NC; Section~\ref{sssec:navie-concatenation}), multichannel (MC; Section~\ref{sssec:multichannel}), separate convolution (SC; Section~\ref{sssec:separate-convolution}), and the three integration approaches with embedding attention ($\ast$-EAV; Section~\ref{ssec:embedding-attention}).
The comparisons of our proposed models to the previous state-of-the-art approaches are outlined in Table~\ref{tbl:resultcomparison}.
For all experiments, the fixed random seed of 1 is used to avoid performance boost from different randomness (see Section~\ref{sssec:random-seed} for more discussions).
The following configuration are used for all models:

\begin{itemize}
\item Filter size = (2, 3, 4, 5) for both word and lexicon embeddings. 
\item Number of filters = (64 and 9) for word and lexicon embeddings, respectively. 
\item Number of filters = (50 and 20) for constructing embedding attention vectors in word and lexicon embedding spaces, respectively. 
\end{itemize}

\noindent It is worth mentioning that the performance of our baseline models improved quite a bit when the training corpora for word embeddings and sentiment analysis were tokenized coherently.
Unlike several other work, we used the identical tokenization tool, NLP4J, to preprocess all corpora, which gave considerable boost in performance.
Comparing the baseline to SC, lexicon embeddings significantly improved accuracy for S16, about 2\%, surpassing the previous state-of-the-art result achieved by \newcite{deriu2016swisscheese}.
However, SC did not show much improvement for SST where the baseline was already performing well.

\noindent Comparing these lexicon integrated models with the ones with embedding attention vectors ($\ast$-EAV), EAV did not help much for S16 but significantly improved the performance for SST, achieving the state-of-the-art result of 48.8\% for a single-layer CNN model.
The accuracy achieved by our best model is still 0.8\% lower than the state-of-the-art result achieved by \newcite{yin2016multichannel}; however, considering their model uses five embedding channels and dual-layer convolutions whereas our model uses a single channel and a single-layer convolution, in other words, our model is much more compact, this is very promising.
These results suggest that lexicon embeddings coupled with the embedding attention vectors allow to build robust sentiment analysis models.

\begin{figure}[!ht]
\centering     
\subfigure[SemEval Task]
{\label{fig:stability_tw}\includegraphics[width=\columnwidth]{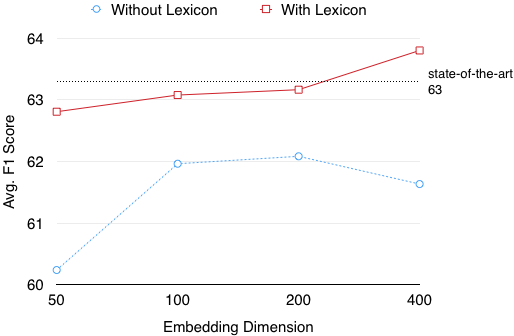}}
\subfigure[SST Task]
{\label{fig:stability_rt}\includegraphics[width=\columnwidth]{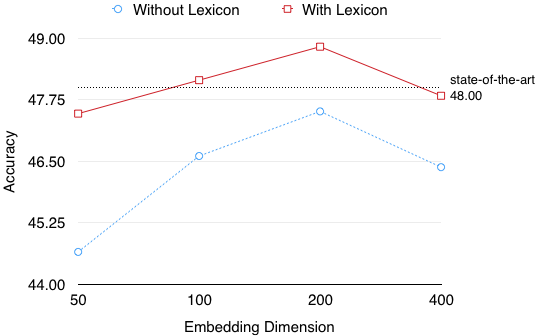}}
\vspace{-2em}
\caption{Performance changes across various dimensions of word embeddings.
}
\label{fig:stability}
\end{figure}

Figure~\ref{fig:stability} illustrates the robustness of our lexicon integrated models with respect to the size of word embeddings.
Our baseline produces inconsistent and unstable results as different sizes of word embeddings are used.
Furthermore, a larger size of word embeddings tends to significantly outperform a smaller size of word embeddings. 
Such tendency is reduced with the incorporation of lexicon embeddings.
While the standard deviations for the accuracies achieved by the baseline using different sizes of word embeddings are 0.8491 and 1.1909 for S16 and SST, respectively,
they are reduced to 0.4208 and 0.5764 respectively for lexicon integrated models.
Furthermore, the accuracy achieved by the lexicon integrated model using the word embedding size 50 is higher or equal to the highest accuracy achieved by the baseline using the word embedding size 200, which implies that it is possible to build more compact models using lexicon embeddings without compromising accuracy.




\subsection{Analysis}

\subsubsection{Randomness in Deep Learning}
\label{sssec:random-seed}

\begin{figure}[!htp]
\vspace{-2ex}
\centering     
\subfigure[SemEval Task: 
The baseline model has a higher variance than the proposed models.
Adding lexicon information improves the baseline model to be more accurate.
In addition, EAV marginally pushes the performance.]
{\label{fig:modelvar_tw}\includegraphics[width=\columnwidth]{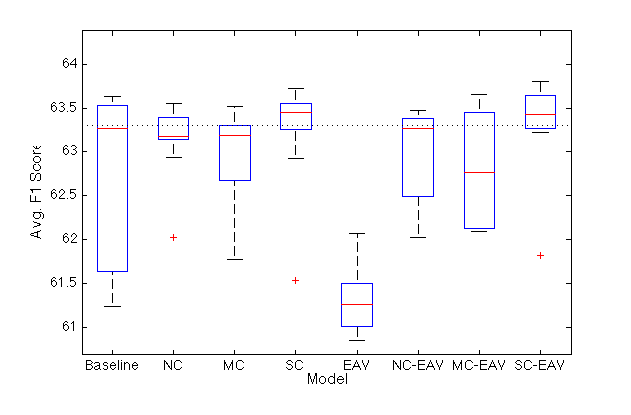}}
\subfigure[SST Task:
The baseline model itself is stable because 
the vocabulary of the word embedding covers 
approximately all words in SST, as shown in Table
\ref{tbl:summaryStatisticEmbedding}.
Although adding lexicon information 
destabilize the model lightly,
lexicon information enhance the accuracy.
EAV is advantageous in general.
This effect is visually shown in this figure, 
when comparing naive concatenation (NC; (Section~\ref{sssec:navie-concatenation}) 
with NC-EAV. 
]
{\label{fig:modelvar_rt}\includegraphics[width=\columnwidth]{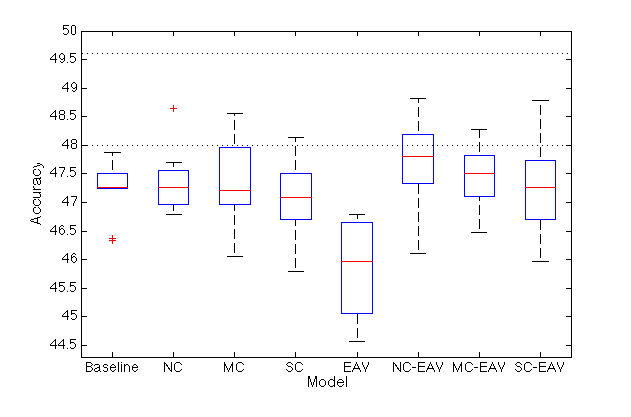}}
\caption{Generalized performance evaluation of the models. 
Each model is trained 10 times with different random seeds and the results are summarized as a bar plot.
In this plot, the central red line indicates the median, 
and the bottom and top edges of the box indicate the 25th and 75th percentiles, respectively. the '+' symbol represents outliers.
}
\label{fig:modelvar}
\end{figure}

Different random seeds when training the CNN models could possibly change the behavior of models, sometimes by more than 1\%. This is due to the randomness in deep learning,
such as the random shuffling the datasets, 
initialization of the weights and
drop-out rate of a tensor.
To reduce the impact of random seed on our result and capture the general characteristic of the model, we performed a group analysis by training each model with 10 different random seeds (Figure~\ref{fig:modelvar}).

\subsubsection{S16: SemEval'16 Task 4}

For S16, the lexicon integration tends to reduce
the variances, and the introducing embedding attention vectors pushes the accuracy even higher 
than the models without it across different random seeds.
Another notable observation for S16
is that although multichannel method underperforms 
when the random seed is fixed to a specific number as seen in
Table~\ref{tbl:resultcomparison}, 
it produces a competitive output in the group analysis setting.
Such low performance with a fixed random seed
is probably attributed to the well known problem of optimization, 
trapping in local optima.


\subsubsection{SST: Stanford Sentiment Treebank}

The problem conditions for SST are different in terms of vocabulary coverage.
This difference is caused by the source of the lexicon embeddings,
where all of them were constructed from Twitter dataset. 
Since most of the lexical words are from Twitter, 
it shows less vocabulary coverage on SST than that of S16
as shown in the right columns of Table~\ref{tbl:summaryStatisticEmbedding}.
Because of this poor relatedness between lexicons and datasets, 
we hypothesized 
that adding a lexicon might be less effective on the performance of SST task.
However, 
our models seems to successfully adopt exogenous features 
enough to push the accuracy marginally higher than the models without lexicons.

On the contrary, the coverage of word embeddings on SST is notably high at around 98\%,
while only around 70\% for S16 (left columns of Table~\ref{tbl:summaryStatisticEmbedding}).
These conditions are well reflected in the group analysis of the model in SST.
Since word embeddings themselves are sufficient enough to 
cover majority of words,
the model variance of the baseline is relatively small compared to S16.


\begin{figure*}[htb]
\centering
\includegraphics[width=\textwidth]{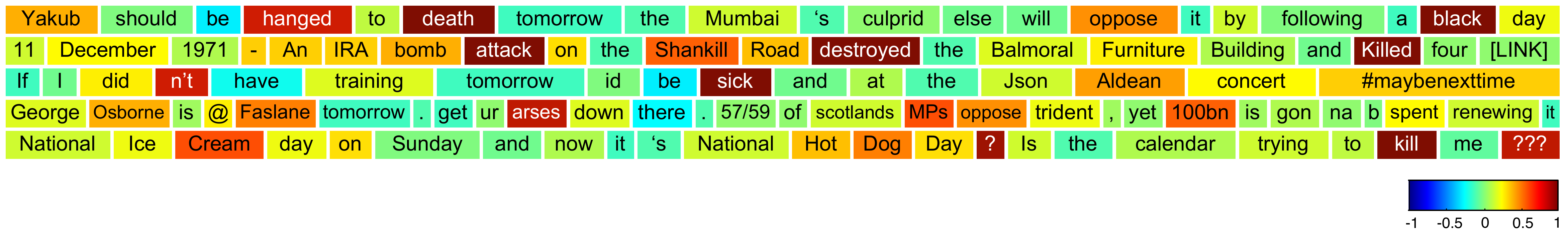}
\caption{Five selected negative tweets with the attention heatmap. 
Examples are from the set where the baseline gives wrong answers
but SC-EAV predicts correctly. 
Intensity of each word roughly ranges from -1 to 1.
This weights (intensities) are 
the values of the attention vector of the word embeddings in the SC-EAV model.
While negative words get more attention (reds),
non-sentimental words such as stop words get less attention (greens and light blues).
}
\label{fig:att}
\end{figure*}

\subsubsection{Attention}

Embedding attention vectors allow to visualize the importance of each word and lexicon for sentiment analysis through a heatmap.
In Figure~\ref{fig:att}, all negative words get higher weights (reds), while non-sentimental words do not (greens and light blues) in EAV.
This visualization is especially useful for neural models because it provides an compelling explanatory information about how the models work.

\subsubsection{Learning Speed}
\begin{figure}[!htbp]
\centering
\includegraphics[width=60mm]{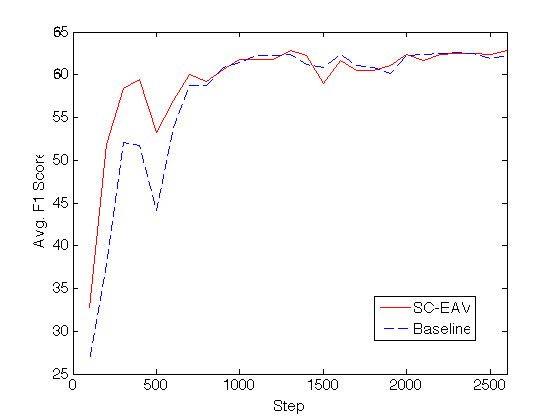}
\caption{Lexicon information and EAV accelerate the learning speed.
High F1 score can be achieved in the earlier step, 
if lexicon information is incorporated along with EAV.}
\label{fig:lc}
\end{figure}

Another advantage of the proposed model, SC-EAV,
is that it accelerates the learning speed (Figure~\ref{fig:lc}).
High F1 score can be achieved in the earlier step,
if lexicon information is incorporated along with EAV.
This statement is general behavior because
the learning curves in Figure Figure~\ref{fig:lc} are the result of averaging  
ten different learning attempts with different random seeds.

\section{Conclusion}
\label{sec:conclusion}

This paper proposes several approaches that effectively integrate lexicon embeddings and an attention mechanism to a well-explored deep learning framework, Convolutional Neural Networks, for sentiment analysis.
Our experiments show that lexicon integration can improve accuracy, stability, and efficiency of the traditional CNN model. 
Multiple training results with different random seeds show the generalization of the effectiveness of using lexicon embeddings and embedding attention vectors. 
The training curve comparison further shows another benefit of this integration for more robust learning.
The attention heatmap analysis confirms that embedding attention vectors endow CNN models with explanatory features, which gives good understanding of how the CNN models work.

\noindent Much more future work is left.
The proposed attention models are applied to each single word. However, focusing on multiple words could give more promising information.
Application of the attention mechanism to multiple words at the same time is a possible direction.
Majority of the lexicons in this work are from tweet dataset. 
More lexicon dataset from general could be used to improve the coverage of our system.
We focused on a simple and yet well performing system.
In order to maximize the score, ensemble of multi layer CNN models could be applied.\footnote{All our resources are publicly available\\: \url{http://nlp.mathcs.emory.edu}}

\section*{Acknowledgments}

We gratefully acknowledge the support of the University Research Committee Grant (URC) from Emory University, and the Infosys Research Enhancement Grant. Any contents
expressed in this material are those of the authors
and do not necessarily reflect the views of these
awards and grants. Special thanks are due to Jung-Hyun Kang for producing the wonderful figures.

\bibliography{wassa2017}
\bibliographystyle{emnlp_natbib}

\end{document}